%% file: main.tex
\documentclass{article}

\usepackage{PRIMEarxiv}

\usepackage{graphicx}
\usepackage{listings}
\usepackage{subcaption}

\usepackage{amsmath}
\usepackage{amsfonts}
\usepackage{amssymb}
\usepackage{algorithm}
\usepackage{algpseudocode}
\usepackage{float}
\usepackage{caption}
\usepackage{url}
\usepackage[square,numbers]{natbib}
\usepackage{xcolor}
\usepackage{colortbl}
\usepackage{orcidlink}
\newcommand{\simplex}{\Delta^{n-1}}

\newcommand{\repetitions}{1000}
\newcommand{\samplesize}{100}
\newcommand{\oleap}{O-LEAP$_\mathrm{KDEy}$}

\newcommand{\haty}{\hat{Y}}
\newcommand{\tpr}{\mathrm{tpr}}
\newcommand{\tnr}{\mathrm{tnr}}
\newcommand{\Ltr}{L_{\mathrm{tr}}}
\newcommand{\Lva}{L_{\mathrm{va}}}

\newcommand{\tmsall}{TMS-All}
\newcommand{\imsall}{IMS-All}
\newcommand{\imslr}{IMS-LR}
\newcommand{\imsknn}{IMS-$k$NN}
\newcommand{\imssvm}{IMS-SVM}
\newcommand{\imsmlp}{IMS-MLP}
\newcommand{\nomslr}{$\varnothing$-LR}
\newcommand{\nomsknn}{$\varnothing$-$k$NN}
\newcommand{\nomssvm}{$\varnothing$-SVM}
\newcommand{\nomstsvm}{$\varnothing$-TSVM}
\newcommand{\nomsmlp}{$\varnothing$-MLP}

\newcommand{\codecomment}[1]{\State {\leavevmode\color{gray}{// {#1}}}}

\DeclareMathOperator*{\argmax}{arg\,max}

\newif\ifdraft
\title{Transductive Model Selection under Prior Probability Shift}

\author{
    Lorenzo Volpi \orcidlink{0009-0006-0851-8041}, Alejandro Moreo \orcidlink{0000-0002-0377-1025}, Fabrizio Sebastiani \orcidlink{0000-0003-4221-6427}\\
    Istituto di Scienza e Tecnologie dell'Informazione,  
    Consiglio Nazionale delle Ricerche\\
    Via Giuseppe Moruzzi 1, 56124, Pisa, Italy\\
  \texttt{\{firstname.lastname\}@isti.cnr.it} \\
}

\begin{document}

\maketitle

\begin{abstract}
 Transductive learning is a supervised machine learning task in
 which, unlike in traditional inductive learning, the unlabelled data
 that require labelling are a finite set and are available at
 training time. Similarly to inductive learning contexts,
 transductive learning contexts may be affected by dataset shift,
 i.e., may be such that the IID assumption does not hold. We here
 propose a method, tailored to transductive classification contexts,
 for performing model selection (i.e., hyperparameter optimisation)
 when the data exhibit prior probability shift, an important type of
 dataset shift typical of anti-causal learning problems. In our
 proposed method the hyperparameters can be optimised directly on the
 unlabelled data to which the trained classifier must be applied;
 this is unlike traditional model selection methods, that are based
 on performing cross-validation on the labelled training data. We
 provide experimental results that show the benefits brought about by
 our method.
\end{abstract}


% ------------------------------------------------------

\section{Introduction}
\label{sec:intro}
\noindent Consider the outbreak of an epidemic, in which the
prevalence of individuals affected by an infectious disease rapidly
increases while the distribution of the symptoms (i.e., the effects)
across the affected individuals remains unchanged. In such a scenario,
we may want to train a classifier that, from the symptoms an
individual displays, infers whether she is affected or not from the
disease. We consider the situation in which (i) the data to be
classified arrives in successive batches $u_1$, $u_2$, \ldots, and,
(ii) given the epidemic, we may expect the prevalence of the affected
individuals to evolve rapidly, i.e., exhibit different values in
different batches. Assume that, for training the classifier, and in
particular for selecting the combination of hyperparameters expected
to yield the best accuracy under the new (i.e., epidemic) conditions,
we have access to training data collected in the old (i.e.,
pre-epidemic) conditions.

This scenario is problematic, as the training data and the unlabelled
data to be classified are not \textit{identically and independently
distributed} (IID), due to the fact that the prevalence of the
individuals affected by the disease has changed from the training data
to the unlabelled data.
% (When the data are not IID, we say that there is \textit{dataset
% shift} between the training data and the unlabelled
% data~\cite{Quinonero:2009kl, Moreno-Torres:2012ay, Storkey:2009lp}.)
The classifier, and the chosen hyperparameter combination, may thus
reveal suboptimal once used on the unlabelled data.

In this paper we present a method for optimising the hyperparameters
(i.e., for performing \textit{model selection} -- MS) \textit{directly
on the batch of unlabelled data that need to be classified}. For
reasons that will be explained in Section~\ref{sec:method}, we call
this task \textit{transductive model selection} (TMS). A TMS method
has obvious advantages over the standard inductive model selection
(IMS) method (that relies on cross-validation on the training data),
since the chosen hyperparameters are tailored to the batch $U_i$ of
unlabelled data that need to be classified, and can thus deliver
better performance on $U_i$ than hyperparameters chosen via standard
IMS.
Our proposed method is based on techniques for \textit{classifier
accuracy prediction} (CAP) \textit{under dataset
shift}~\cite{Garg:2022qv, Guillory:2021so, Volpi:2024ye}, and
essentially consists of (i) predicting the accuracy that different
classifiers, instantiated with different choices of hyperparameters,
would obtain on our ``out-of-distribution'' batch $U_i$, and (ii)
classifying the data in batch $U_i$ using the classifier whose
predicted accuracy is highest. In other words, our TMS method replaces
traditional accuracy \textit{computation} on \textit{labelled} data
with accuracy \textit{estimation} on \textit{unlabelled} data. While
the method is generic, we here restrict our attention to the case in
which the data are affected by \textit{prior probability shift} (PPS),
an important
% type of dataset shift.
type of violation of the IID assumption.

For high-stakes applications such as the healthcare-related
classification task discussed above, this suggests a policy of (a)
training, on the labelled data, multiple classifiers, each
characterised by a different combination of
hyperparameters,\footnote{Note that this step is performed anyway
during traditional hyperparameter optimisation.} (b) storing these
classifiers for later use, and, every time a new batch $U_i$ of
unlabelled data becomes available, (c) estimating (via CAP techniques)
the accuracy that the different classifiers would have on $U_i$ and
(d) classifying the data in $U_i$ via the classifier whose estimated
accuracy is highest. In this way, assuming that Step (b) can be
carried out efficiently, the classification of newly arrived data can
be performed immediately and, as we will show, with a much higher
accuracy than can be obtained via the traditional model selection
method. Note that Step (a) is carried out only once, since we do not
assume new labelled data to become available during the process.

The rest of the paper is organised as
follows. Section~\ref{sec:method} introduces the notation and provides
a detailed description of the proposed
method. Section~\ref{sec:experiments} presents the experiments we have
carried out and discusses the results we have
obtained. Section~\ref{sec:conclusions} concludes the paper with a
summary of our findings and a discussion of potential applications of
this method.

% ------------------------------------------------------

\section{Transductive Model Selection under Prior Probability Shift}
\label{sec:method}

% In machine learning, classification tasks are typically tackled by
% experimenting with many classifiers and many configurations among
% them, evaluating their performances and choosing the best. In real
% world scenarios, this can be difficult since finding a classifier
% that generalises well for every domain of application can be
% daunting. It is true that, as long as the training distribution (the
% distribution on which the classifier was trained) and the target
% distribution (the one on which the classifier will be applied) are
% Identically and Independently Distributed (IID), good generalisation
% results can still be achieved with common techniques such as Cross
% Validation and the Holdout Method. Unfortunately, in real word
% applications the training and target distributions are often
% characterised by \emph{dataset shift}, i.e., the IID assumption does
% not hold. In such cases, choosing the best classifier or selecting
% the best hyperparameters for a classifier of choice on the training
% distribution can generate poor results, as the IID assumption that
% leads the choice is not valid.
% 
% To overcome this issue, one would need to have access to the labels
% of at least a portion of the target distribution. Unfortunately,
% this is rarely an option, because either such labels are not
% available or it is too costly to generate them (e.g., via human
% intervention).

% In machine learning, it is common practice to instantiate multiple
% models with different hyperparameters when addressing a
% classification task.
\noindent In supervised machine learning we use training data to learn
the internal parameters of the model (e.g., the weights of a neural
network, or the coefficients of a hyperplane in an SVM). Many models
trained in this way also rely on a set of hyperparameters (e.g., the
learning rate in neural training, or the trade-off between margin and
training error in SVMs) that impose higher-level constraints on the
learning process. Unlike the internal parameters of the model,
hyperparameters are not learned during training, but must be set in
advance. Finding good values for the hyperparameters is crucial for
achieving good performance.
% Selecting the best-performing model generally involves evaluating
% their performance employing standard model selection techniques,
% such as three-way hold out or $k$-fold
% cross-validation~\citep{Raschka2018ms}.
Model selection, the task of choosing the values of the
hyperparameters, is typically carried out by (a) testing the accuracy
of the model under different combinations of hyperparameter values
using cross-validation on labelled data, and (b) choosing the
combination of hyperparameter values that maximizes model
accuracy. %~\citep{Raschka2018ms}.

% However, in real-world scenarios, this process becomes more
% challenging, as finding a classifier that generalises well across
% all cases can be difficult. These standard approaches rely on the
% assumption that the training and test datasets are drawn from the
% same underlying distribution, i.e., they are \emph{independently and
% identically distributed} (IID).
Relying on labelled data to evaluate differently configured models
requires the labelled data to be representative of the unlabelled data
the trained model will be applied to, a distributional assumption
typically referred to as the IID assumption.
% In practice, this assumption is often violated, and the
% distributions may differ, resulting in what is known as
% \emph{distribution shift}~\cite{Storkey:2009lp}.
Unfortunately, in real-world problems this assumption is often
violated; in this case, the training data are \textit{not}
representative of the unlabelled data (which are thus said to be
``out-of-distribution'' data), and we say that the problem is affected
by \emph{dataset shift}~\cite{Storkey:2009lp}.
% Due to the presence of distribution shift, adopting conventional
% model selection techniques to choose the most accurate classifier
% can lead to suboptimal or even misleading results. A model selected
% based on its performance on the training distribution may not
% exhibit the same level of performance when deployed on the test
% distribution.
In problems characterised by dataset shift, cross-validation on
training data is thus a biased estimator of model
accuracy~\citep{SugiyamaM05}, and this often leads to suboptimal
choices of the hyperparameter values.
In classification, one type of dataset shift of particular relevance
is \emph{prior probability shift} (PPS)~\cite{Storkey:2009lp}, also
known as \emph{label shift}~\citep{Lipton:2018fj}. This type of shift
(sometimes considered the ``paradigmatic'' case of dataset shift in
classification~\cite{Ziegler:2024qq}) is characteristic of
\textit{anti-causal learning} problems~\citep{Scholkopf:2012je} (also
known as $Y\rightarrow X$ \textit{problems}~\citep{Fawcett:2005fk},
where $Y$ is a random variable ranging on the class labels and $X$ is
a random variable ranging on vectors of covariates), i.e., problems
where the goal is to predict the causes of a phenomenon from its
observed effects.

PPS is characterised by two distributional assumptions, often called
the \textit{PPS assumptions}, i.e.,
\begin{enumerate}
\item the class priors of the training distribution differ from those
 of the distribution of the unlabelled data (in symbols: $P(Y) \neq
 Q(Y)$, where $P$ and $Q$ are the distributions from which the
 training data and the unlabelled data are sampled, respectively);
\item the class-conditional distribution of the covariates in $P$ is
 the same as that in $Q$ (in symbols: $P(X|Y)=Q(X|Y)$).
\end{enumerate}
\noindent The healthcare-related problem discussed in
Section~\ref{sec:intro} is indeed an anti-causal learning
problem. Indeed, if we take random variable $Y$ to range over
$\mathcal{Y}=\{\textsf{Disease},\textsf{NoDisease}\}$, and random
variable $X$ to range over the vectors of covariates representing
symptoms exhibited by individuals, the anti-causal nature of the
problem is evident. If we take $P$ and $Q$ to be the data
distributions characterizing the pre-epidemic and the epidemic
scenarios, respectively, we are in the presence of PPS, since $P(Y)
\neq Q(Y)$ (the prevalence values of \textsf{Disease} and
\textsf{NoDisease} have changed when switching from $P$ to $Q$) and
$P(X|Y)=Q(X|Y)$ (the distributions of the symptoms exhibited by
affected individuals are the same in $P$ and $Q$).

In the presence of PPS (as in the presence of any other type of shift,
for that matter), a classifier whose hyperparameters have been
optimised on data from $P$ may behave suboptimally when applied to
data from $Q$ (see Section~\ref{sec:trust} for a formal proof). For it
to behave optimally on data from $Q$, hyperparameter optimisation
should have been carried out on data from $Q$, but this is not
possible if using standard cross-validation techniques, since the
labels of data from $Q$ are not known.

To address this problem, we introduce \emph{transductive model
selection} (TMS), a new strategy aimed at selecting the hyperparameter
configuration for a given classifier (or the model from a pool of
already trained candidates) that is predicted to be the best for a
specific batch $U_i$ of unlabelled data characterised by dataset
shift.
% Traditional model selection approaches, such as cross-validation,
% become unreliable under dataset shift, making it challenging to
% evaluate competing classifiers on the shifted test set. Although
% some prior work has addressed model selection under covariate
% shift~\citep{SugiyamaM05}, the problem remains largely unexplored in
% the context of prior probability shift (PPS), which is the focus of
% our study.
This strategy leverages recent advances in classifier accuracy
prediction (CAP)~\cite{garg2022leveraging, Guillory:2021so,
Volpi:2024ye} (a family of techniques specifically designed to
estimate classifier accuracy under dataset shift), and focuses in
particular on CAP methods tailored to PPS~\citep{Volpi:2024ye}.

% \alexcomment{Questo lo toglierei; l'ho già incluso sopra} Although
% dataset shift is a widespread issue in applied machine learning, the
% topic of model selection under dataset shift has received relatively
% little attention in recent years. While some studies have addressed
% model selection under different types of shift, such as the work
% by~\citet{SugiyamaM05} on model selection under covariate shift, the
% problem remains largely unexplored in the context of PPS.

% ------------------------------------------------------

\subsection{Why Can't We Trust Cross-Validation Estimates under Prior
Probability Shift?}
\label{sec:trust}
\noindent Assume our classifier accuracy measure is (``vanilla'')
accuracy, i.e., the fraction of classification decisions that are
correct, and assume we have computed an (arbitrarily good) estimate of
the classifier's future accuracy by means of cross-validation on
training data. As training data is drawn from distribution $P$, our
estimate is an approximation of $P(\hat{Y}=Y)$, where $\hat{Y}$ is a
random variable ranging on the predicted class labels. Assume our
unlabelled data is drawn from a distribution $Q$ related to $P$ via
PPS: can we trust our estimate? For simplicity, let us focus on a
generic binary problem (with $\mathcal{Y}=\{0,1\}$) Note that
% Let us begin by noting that our estimate of classifier accuracy
% approximates $P(\hat{Y}=Y)$; note that:
%
\begin{align}
 \begin{split}
 \label{eq:lawofttotalprob1}
 P(\haty=Y) & = P(\haty=1,Y=1)+P(\haty=0,Y=0) \\
 & = P(\haty=1| Y=1)P(Y=1)+P(\haty=0|Y=0)P(Y=0)
 \end{split}
\end{align}
\noindent and that, similarly,
\begin{align}
 \label{eq:lawofttotalprob2}
 Q(\haty=Y) 
 & = Q(\haty=1| Y=1)Q(Y=1)+Q(\haty=0|Y=0)Q(Y=0)
\end{align}
\noindent Our goal is to verify whether $P(\hat{Y}=Y) = Q(\hat{Y}=Y)$
under PPS. We first observe that, as shown in~\cite[Lemma
1]{Lipton:2018fj}, the PPS assumption $P(X|Y)=Q(X|Y)$ (see
Section~\ref{sec:method}) implies that $P(f(X)|Y)=Q(f(X)|Y)$ for any
deterministic and measurable function $f$. In particular, if we take
$f$ to be our classifier $h$, it holds that
$P(\haty|Y)=Q(\haty|Y)$. This means that $P(\haty=1| Y=1)$ and
$Q(\haty=1| Y=1)$ are equal; we indicate both by the symbol ``tpr'',
since they both represent the \emph{true positive rate} of the
classifier. Similarly, $P(\haty=0| Y=0)$ and $Q(\haty=0| Y=0)$ are
equal, and we indicate both as ``tnr'', which stands for the
\emph{true negative rate} of the classifier. We can further simplify
our equations via the shorthands $p=P(Y=1)$ and $q=Q(Y=1)$. Now, let
us assume that $P(\haty=Y)=Q(\haty=Y)$ holds; in this case,
Equations~\ref{eq:lawofttotalprob1} and~\ref{eq:lawofttotalprob2} lead
us to
\begin{align}
 \begin{split}
 \label{eq:derivations}
 P(\haty=Y)&=Q(\haty=Y) \\
 \tpr\cdot p + \tnr \cdot (1-p) &= \tpr\cdot q + \tnr \cdot (1-q) \\
 p \cdot (\tpr-\tnr) + \tnr &= q \cdot (\tpr-\tnr) + \tnr %\\
 % p &= q
 \end{split}
\end{align}
\noindent PPS means that $p\neq q$; Equation~\ref{eq:derivations} thus
implies that $P(\haty=Y)=Q(\haty=Y)$ holds only if $(\tpr-\tnr)=0$.
% or when $p=q$ (i.e., when $P(Y=1)=Q(Y=1)$). The second condition
% directly contradicts one of the assumptions of PPS (that is,
% $P(Y)\neq Q(Y)$), meaning that our estimate of $P(\haty=Y)$ cannot
% be a good estimate for $Q(\haty=Y)$ unless $\tpr=\tnr$.
However, $\tpr=\tnr$ is not true in general (and is unlikely to be
true in practice); assuming that $P(\haty=Y)=Q(\haty=Y)$ thus leads to
a contradiction. A similar reasoning holds for the multiclass case.

% ------------------------------------------------------

\subsection{Model Selection: From Induction to Transduction}

\noindent Let us assume the following problem setting. Let $\Theta$ be
the set of all assignments of values to hyperparameters that we want
to explore as part of our model selection process; in this paper we
will concentrate on a standard \emph{grid search} exploration,
although other strategies (e.g., Gaussian processes, randomized
search) might be used instead. Let $L$ be our (\underline{l}abelled)
training set and $U_i$ our batch of (\underline{u}nlabelled)
data. Consider the class $\mathcal{H}$ of hypotheses, and let
$h^\theta\in\mathcal{H}$ be the classifier with hyperparameters
$\theta$ trained via some learning algorithm $A$ using labelled data
$L$. Let $M : \mathcal{H}\times \mathcal{U}\rightarrow \mathbb{R}$ be
the measure of accuracy for a classifier $h\in\mathcal{H}$ on batch
$U_i\in\mathcal{U}$ of unlabelled data we want to optimise $h$
for. The model selection problem can thus be formalized as
\begin{equation}
 \theta^*=\argmax_{\theta\in\Theta} M(h^\theta, U_i)
\end{equation}
\noindent Since we do not have access to the labels in $U_i$, the
problem cannot be solved directly, and we must instead resort to
approximations. The most common way for solving it corresponds to the
traditional \emph{inductive model selection} method (IMS --
Section~\ref{sec:ims}).

% ------------------------------------------------------

\subsubsection{Inductive Model Selection}
\label{sec:ims}

\noindent The IMS approach comes down to using part of the training
data for evaluating each configuration of hyperparameters, based on
the assumption that such an estimate of classifier accuracy
generalizes for future data. In this paper we carry this out via
standard cross-validation (although everything we say applies to
$k$-fold cross-validation as well), splitting $L$ (with
stratification) into a proper training set $\Ltr$ and a validation set
$\Lva$.
% IMS can be described as:
IMS is described in Algorithm~\ref{alg:ims}.

However, since IMS is unreliable under PPS for the reasons discussed
in Section~ \ref{sec:trust}, we propose an alternative model selection
method called \emph{transductive model selection} (TMS --
Section~\ref{sec:tms}).
% \alexcomment{Riusciamo a far vedere i due pseudocodici uno accanto
% all'altro?}

% \begin{algorithm}
% \caption{Inductive Model Selection}\label{alg:ims}
% \begin{algorithmic}
%% \footnotesize \Input $\Ltr, \Lva$
% \For{$\theta \in \Theta$} \State $h_\theta\leftarrow
% A(\mathcal{H},\theta,\Ltr)$ \State
% $\widehat{\mathrm{Acc}}\leftarrow M(h_\theta, \Lva)$
% \If{$\widehat{\mathrm{Acc}} > \mathrm{BestAcc}$} \State
% $\mathrm{BestAcc} \gets \widehat{\mathrm{Acc}}$ \State
% $h^*_\theta \gets h_\theta $ \EndIf \EndFor
% \\
% \Return $h^*_\theta$ \Comment{Returns an inductive classifier
% that can be applied to any test set}
% \end{algorithmic}
% \end{algorithm}

% ------------------------------------------------------

\subsubsection{Transductive Model Selection}
\label{sec:tms}

\noindent The main difference between IMS and TMS lies in the accuracy
estimation step. Unlike IMS, which estimates the accuracy on
unlabelled data by computing accuracy on labelled (validation) data,
TMS estimates the accuracy on unlabelled data directly on the
available set of unlabelled data. To this aim, TMS employs a
classifier accuracy prediction (CAP) method, i.e., a predictor
$\psi_{h}:\mathcal{U}\rightarrow \mathbb{R}$ of the accuracy that $h$
will exhibit on a batch $U_i\in\mathcal{U}$ of unlabelled
data. However, this does not mean that TMS can avoid using part of the
labelled data, since it still requires a portion of it to train the
CAP method. Since the procedure is transductive, its outcome is not a
generic classifier that can be applied to any future data, but the set
of labels assigned to the unlabelled instances in $U_i$. TMS is
described in Algorithm~\ref{alg:tms}.

% \begin{algorithm}
% \caption{Transductive Model Selection}\label{alg:tms}
% \begin{algorithmic}
%% \REQUIRE $\Ltr, \Lva, U$ \footnotesize
% \For{$\theta \in \Theta$} \State $h_\theta\leftarrow
% A(\mathcal{H},\theta,\Ltr)$ \State $\psi_{h_\theta} \gets
% \mathrm{CAP}(h_\theta, \Lva)$ \State
% $\widehat{\mathrm{Acc}}\gets \psi_{h_\theta}(U)$
% \If{$\widehat{\mathrm{Acc}} > \mathrm{BestAcc}$} \State
% $\mathrm{BestAcc} \gets \widehat{\mathrm{Acc}}$ \State
% $h^*_\theta \gets h_\theta $ \EndIf \EndFor
% \\
% \Return $\{(x_i,h^*_\theta(x_i)):x_i\in U\}$ \Comment{Returns
% the transduced labels for the specific test set}
% \end{algorithmic}
% \end{algorithm}

% \alexcomment{Provo a metterli uno accanto all'altro...}
\begin{figure}[th]
 \centering
 \begin{minipage}{0.48\textwidth}
 \captionof{algorithm}{Inductive Model Selection}\label{alg:ims}
 \begin{algorithmic}
 \For{$\theta \in \Theta$} \State $h_\theta \gets
 A(\mathcal{H},\theta,\Ltr)$ \codecomment{Trains the classifier
 via algorithm $A$} \State $\mathrm{Acc} \gets M(h_\theta,
 \Lva)$ %\aftercodecomment{Estimates accuracy on validation}
 \codecomment{Computes accuracy on validation data}
 \If{$\mathrm{Acc} > \mathrm{BestAcc}$} \State $\mathrm{BestAcc}
 \gets \mathrm{Acc}$ \State $h^*_\theta \gets h_\theta $ \EndIf
 \EndFor
 \\
 \Return $h^*_\theta$ %\State $\phantom{.}$
 \codecomment{Returns an inductive classifier that can be }
 \codecomment{applied to any set of unlabelled data}
 \vspace{5.4ex}
 \end{algorithmic}
 \end{minipage}
 \hfill
 \begin{minipage}{0.48\textwidth}
 \captionof{algorithm}{Transductive Model Selection}\label{alg:tms}
 \begin{algorithmic}
 \For{$\theta \in \Theta$} \State $h_\theta\gets
 A(\mathcal{H},\theta,\Ltr)$ \codecomment{Trains classifier
 $h_\theta$ via algorithm $A$} \State $\psi_{h_\theta} \gets
 \mathrm{CAP}(h_\theta, \Lva)$ \codecomment{Trains a CAP method
 for classifier $h_\theta$} \State $\widehat{\mathrm{Acc}}\gets
 \psi_{h_\theta}(U_i)$ % \aftercodecomment{Estimates accuracy on the test set}
 \codecomment{Estimates accuracy on the unlabelled data}
 \If{$\widehat{\mathrm{Acc}} > \mathrm{BestAcc}$} \State
 $\mathrm{BestAcc} \gets \widehat{\mathrm{Acc}}$ \State
 $h^*_\theta \gets h_\theta $ \EndIf \EndFor
 \\
 \Return $\{(x_j,h^*_\theta(x_j)):x_j\in U_i\}$
 \codecomment{Returns the inferred labels for the specific}
 \codecomment{unlabelled data}
 \end{algorithmic}
 \end{minipage}
\end{figure}

\section{Experiments}
\label{sec:experiments}

\noindent In this section we present an experimental comparison
between IMS and TMS under PPS.\footnote{The code to reproduce all our
experiments is available on GitHub at
\url{https://github.com/lorenzovolpi/tms}}
% conducted to demonstrate the effectiveness of LEAP as a CAP method
% applied to transductive model selection, and we discuss the results
% obtained by comparing them with those of standard, or na\"ive, model
% selection methods.

% ------------------------------------------------------

\paragraph*{Experimental Protocol.}
\label{sec:exp_setup}

\noindent The effectiveness measure we use in order to assess the
quality of the model selection strategies is the (``vanilla'')
accuracy of the selected model, i.e., the fraction of classification
decisions that are correct.
% model selection methods is the true accuracy obtained by the
% selected classifier on the test set. As our classifier accuracy
% measure we here employ vanilla accuracy, one of the most important
% measures of classifier effectiveness.

% For each dataset $D=\{(\x_i,y_i)\}_{i=1}^N$

The experimental protocol we adopt is as follows. Given a dataset $D$,
we split it into a training set $L$ (70\%) and a test set $U$ (30\%)
via stratified sampling; we further split the training set into a
``proper'' training set $\Ltr$ and a validation set $\Lva$, with
$|\Ltr|=|\Lva|$, via stratification.

In order to simulate PPS we apply the Artificial Prevalence Protocol
(APP)~\cite{Esuli:2023os, Forman:2008kx}
% \alexcomment{cita}
on $U$. This consists of drawing $r$ vectors $\mathbf{v}_1, ...,
\mathbf{v}_{r}$ (we here take $r=\repetitions$) of $n$ prevalence
values (with $n$ the number of classes) from the unit simplex
$\simplex$ (using the Kraemer sampling
algorithm~\citep{smith2004sampling}), and extracting from $U$, for
each $\mathbf{v}_i$, a bag $U_i$ of $|U_i|=s$ elements (we here take
$s=\samplesize$) such that $U_i$ satisfies the prevalence distribution
of $\mathbf{v}_i$.\footnote{We use the term ``bag'' (i.e., multiset)
since we sample with replacement, which might lead to $U_i$ containing
duplicates.} We then run our experiments by
\begin{enumerate}
\item training all the differently configured classifiers on $\Ltr$;
\item \label{step:compute} for applying IMS: computing the accuracy of
 each trained classifier on $\Lva$ and classifying the datapoints in
 all the $U_i$'s via the classifier that has shown the best accuracy;
\item \label{step:estimate} for applying TMS: for each $U_i$,
 estimating the accuracy on $U_i$ of each trained classifier via a
 CAP method trained on $\Lva$, and applying to $U_i$ the classifier
 that has shown the best accuracy.

\end{enumerate}

\paragraph*{Classifiers and Hyperparameters.}
\label{sec:classifiers}

\noindent We test both model selection approaches on four classifier
types, namely, classifiers trained via Logistic Regression (LR),
$k$-Nearest Neighbours ($k$-NN), Support Vector Machines (SVM), and
Multi-Layered Perceptron (MLP). Each classifier type is instantiated
with multiple combinations of hyperparameters, with the total number
of combinations depending on the number of classes in the dataset and
on the classifier type.
% Note that we do not explore any instantiation of TSVM
% \alexcomment{non lo usiamo?} \lorenzocomment{no, al momento lo
% usiamo solo come baseline; se vogliamo togliere questa parte che può
% creare confusione lo posso far girare (solo TSVM), dovrebbe far
% presto} in our model selection experiments, given that this
% classification model is not supposed to be employed in a context
% dominated by dataset shift (and in particular by PPS, as in our
% experiments).

% For both LR and SVM classifiers, we consider six different values
% for the \verb|class_weight| hyperparameter in the binary
% classification setting. Four of these, are obtained by varying the
% prevalence value of the positive class on a grid
% $(0.2,0.4,0.6,0.8)$; another one uses the \verb|balanced| setting;
% and the last one leaves the class weights at their default
% values. In the multiclass case, the number of \verb|class_weight|
% values depends on the number of classes, specifically amounting to
% $(n + 2)$, where $n$ is the number of classes. $n$ of these are
% generated by constructing $n$ distinct probability distributions: in
% each distribution, one class was assigned a ``high'' weight of
% $2/n$, while the remaining $(n-1)$ classes where assigned a ``low''
% weight of %$\left(1 - \frac{2}{n}\right)\cdot\frac{1}{n-1}$.
% $\left(\frac{1-2/n}{n-1}\right)$. These $n$ distributions are
% created by varying the class receiving the high weight.

% \lorenzocomment{
Under PPS, one of the most interesting hyperparameters is probably the
\texttt{class\_weight} hyperparameter of LR and SVM, which allows
rebalancing the relative importance of the classes to compensate for
class imbalance. In the presence of PPS, exploring different
class-balancing configurations increases the probability of
instantiating a classifier trained according to a class importance
scheme that fits well the unlabelled data. For LR and SVM, we consider
different values for the \texttt{class\_weight} hyperparameter
depending on the number $n$ of classes. In all cases, we include the
configurations \texttt{balanced} (which assigns different weights to
instances of different classes to compensate for class imbalance in
the training data) and \texttt{None} (all instances count the same,
which results in more popular classes dominating the learning
process).
% \fabseb{Value \texttt{balanced} is optimal when there is no PPS, but
% if we assume we might be in the presence of PPS, we cannot
% anticipate how the priors have changed ...}
Aside from these, we explore alternative \texttt{class\_weight} values
that try to compensate potentially high values of a single class, one
class at a time. The reason why we limit ourselves to this kind of
exploration is to prevent combinatorial explosion; focusing on more
than one class at a time, or on more weight values, would result in a
potentially un-manageable number of hyperparameter combinations,
especially for high values of $n$.
% Aside from these, alternative \texttt{class\_weight} values
These alternative values must be specified as points in the
probability simplex (i.e., the per-class balancing weights must add up
to one). In multiclass problems with $n>2$, we add $n$ such
configurations to the pool of values, which we obtain as all different
permutations composed of one ``high'' weight and $(n-1)$ ``low''
weights. We set the high value to $2/n$ (i.e., twice the mass of a
uniform assignment) and distribute the remainder among the low values,
thus setting each to $\left(\frac{1-2/n}{n-1}\right)$.\footnote{
Notice that the choice of $2/n$ as the high value is arbitrary, but we
consider it a good heuristic when trying to compensate for class
imbalance. } For example, when $n=3$ we explore the
\texttt{class\_weight} assignments $(0.66, 0.165, 0.165)$, $(0.165,
0.66, 0.165)$, and $(0.165, 0.165, 0.66)$. In the binary setting,
where a finer-grained set of combinations is manageable, we instead
use a grid of class weights $G$ and explore all combinations $(g,
1-g)$ with $g\in G$. In particular, we use the grid
$G=(0.2,0.4,0.6,0.8)$, thus considering the \texttt{class\_weight}
assignments $(0.2, 0.8)$, $(0.4, 0.6)$, $(0.6, 0.4)$, and $(0.8,
0.2)$.
% When $n=2$ (i.e., in the binary setting), this configuration would
% result in excessively unbalanced weights, so we explore , instead,
% the prevalence value of the positive class on a grid
% $(0.2,0.4,0.6,0.8)$. }

Concerning the other hyperparameters, we consider five different
values ($10^{-2}$, $10^{-1}$, $10^{0}$, $10^{1}$, $10^{2}$) for
hyperparameter \verb|C| (the regularization strength for both LR and
SVM), as well as two additional values of \verb|gamma| (\verb|scale|
and \verb|auto|) for SVM only. For $k$-NN we explore five values of
$k$, i.e., of \verb|n_neighbors| (5, 7, 9, 11, 13) and two values of
\verb|weights| (\verb|uniform| and \verb|distance|). For MLP, we test
five values of \verb|alpha| ($10^{-5}$, $10^{-4}$, $10^{-3}$,
$10^{-2}$, $10^{-1}$) and two values of \verb|learning_rate|
(\verb|constant| and \verb|adaptive|). We use all these configurations
consistently across all
% both the binary and multiclass
our experiments.\footnote{Hyperparameter names, as well as their
default configurations discussed in the next paragraphs, are those
provided by the \emph{scikit-learn} Python library
(\url{https://scikit-learn.org/}).
% ~\citep{Pedregosa:2011yo}
}

% The total amount of possible hperparameter combinations amounts to
% $30$ for LR in the binary setting and $(n+2) \cdot 5$ in the
% multiclass setting; $60$ for SVM in the binary case and $(n+2) \cdot
% 10$ in the multiclass case; and $10$ for both MLP and $k$-NN in both
% the binary and multiclass settings of our experiments.

% ------------------------------------------------------

\paragraph*{Datasets.}
\label{sec:datasets}

\noindent We use the 25 datasets from the UCI machine learning
repository\footnote{\url{https://archive.ics.uci.edu/}} that can be
directly imported through UCI's Python API and that have at least 6
features and at least 800 instances.
The number of instances per dataset varies from 830
(\textsf{mammographc}) to 1,024,985 (\textsf{poker-hand}), while the
number of features varies from 6 (\textsf{mhr}) to 617
(\textsf{isolet}). %\fabseb{Penso sia necessariao parlare di tutti i
% datasets e non solo dei binary ones rimuovere la distinzione fra
% binary e
% multiclass.} %\lorenzocomment{Ho corretto, il ``binary'' era
% un refuso rimasto dalla vecchia versione del paragrafo.}
The number of classes varies from 2 (\textsf{german},
\textsf{mammographic}, \textsf{semeion}, \textsf{spambase},
\textsf{tictactoe}) to 26 (\textsf{isolet}, \textsf{letter}). Class
balance is highly variable, from datasets in which some of the classes
represent less than 1\% of the instances (e.g., one of the classes in
\textsf{poker-hand}), to perfectly balanced datasets (e.g.,
\textsf{image-seg}).

% ------------------------------------------------------

\paragraph*{Implementation Details for TMS.}
\label{sec:implem}

\noindent As our choice of the CAP method we adopt \oleap, a member of
the \textit{Linear Equations for Accuracy Prediction} (LEAP) family
\cite{Volpi:2024ye} specifically devised for
PPS. %\lorenzocomment{se abbiamo problemi di spazio, possiamo togliere fino al punto ->}
LEAP methods work by estimating the values of the cells of the
contingency table deriving from the application of the classifier to
the set of unlabelled data, where the estimation is obtained by
solving a system of linear equations that represent the problem
constraints (including the PPS assumptions); once the contingency
table is estimated, any classifier accuracy measure can be computed
from it. LEAP internally relies on a \textit{quantifier} (i.e., a
predictor of class prevalence values)~\cite{Esuli:2023os}; following
\cite{Volpi:2024ye}, for this purpose we employ the KDEy-ML
quantification method \cite{Moreo:2025lq}.
% as an intermediate step, enabling the computation of multiple
% evaluation metrics from a single instance. LEAP represents a
% flexible CAP method and candidates as a well-suited choice for TMS.
From now on, we will refer to the TMS method that uses \oleap\ simply
as \tmsall, since this method selects the best model across all
classifier types (LR, SVM, $k$-NN, MLP) and all their hyperparameter
combinations. %(implying that, for this method, hyperparameter
% optimisation is performed considering all classifiers at once).

% ------------------------------------------------------

\paragraph*{Baselines.}
\label{sec:baselines}

\noindent We compare \tmsall\ against standard (inductive) approaches
for model selection. In particular, we consider two variants of IMS:
one in which model selection is performed independently for each
classifier type (\imslr, \imssvm, etc.), and another in which model
selection chooses among all classifier types and all hyperparameter
combinations for each type (\imsall); in other words, \tmsall\ stands
to TMS as \imsall\ stands to IMS. We also compare these model
selection strategies against configurations of each classifier
(\nomslr, \nomssvm, etc.) in which default hyperparameters are used.

We also consider \nomstsvm, an instance of the transductive support
vector machine (TSVM) algorithm
\cite{Gammerman:1998ay}, %\alexcomment{cita:[vapnik, 1995]} \fabseb{Intendi~\cite{Cortes:1995to}?} \lorenzocomment{forse qui è meglio citare~\cite{Gammerman:1998ay}}
which directly infers the label of each unlabelled datapoint without
generating a classifier.\footnote{For TSVM we use the implementation
proposed by \cite{Joachims:1999fr}.
% \fabseb{Intendi~\cite{Joachims2003}?}
}
% \fabseb{Si tratta di un oracolo? Non capisco, ``directly transduces
% the labels of each test set'' è linguaggio da iniziati; in questo
% papero una spiegazione più semplice sarebbe più adatta;
% ``transduces'' vuol dire ``infers''?} \lorenzocomment{Non è un
% oracolo, è la formulazione di Vapnik del Transductive SVM. Comunque
% sì, credo che Alejandro intendesse ``transductively infers'';
% potremmo sostituire la parte ``which directly transduces ...'' con
% $\rightarrow$} \lorenzocomment{ which directly infers the labels of
% each test set $U_i$ adopting a ``particular to particular'' kind of
% inference (or \textit{transductive} inference), rather than
% formulating a general inference rule from the training distribution
% to be then applied to the test samples (i.e., a ``particular to
% general'' type of inference, or \textit{inductive} inference). }
While TSVM relies on the IID assumption and is not a proper model
selection approach, we include it as a reference baseline because it
captures the essence of transductive learning, and thus offers a
meaningful point of comparison. For TSVM, we only consider
instantiations with default hyperparameters since, to the best of our
knowledge, there is no established way to tune hyperparameters for
non-IID settings.

\paragraph*{Results.}
\label{sec:results}

% \begin{table}
% \begin{subtable}[tb]{\textwidth}
% \setlength{\tabcolsep}{2pt} \centering
% \resizebox{\textwidth}{!}{%
% \input{tables/binary_vanilla_accuracy_table} }%
% \caption{Results on the binary datasets.}
% \label{tab:results:bin}
% \end{subtable}
% %
% \begin{subtable}[tb]{\textwidth}
% \setlength{\tabcolsep}{2pt} \centering
% \resizebox{\textwidth}{!}{%
% \input{tables/multiclass_vanilla_accuracy_table} }%
% \caption{Results on the multiclass datasets.}
% \label{tab:results:mul}
% \end{subtable}
% %
% \caption{Results}
% \label{tab:results}
% \end{table}

\begin{table}[tb]
 \setlength{\tabcolsep}{2pt} \centering \resizebox{\textwidth}{!}{%
 \input{tables/transd_vanilla_accuracy_table} }%
 \caption{Classification accuracy results, obtained via classifiers
 whose hyperparameters have been optimised either via IMS 
 % via the standard IMS
 % technique (cross-validation on the labelled data) 
 or via TMS. \textbf{Boldface} represents the best result obtained for
 the given dataset. Superscript $\dag$ denotes the methods (if any)
 whose scores are not statistically significantly different from the
 best one according to a Wilcoxon signed-rank test at 0.01 confidence
 level. Cells are colour-coded so as to facilitate readability, with
 green indicating best and red indicating worst.}
 \label{tab:all_results}
\end{table}

\begin{figure}[tb]
 \centering
 \includegraphics[width=\textwidth]{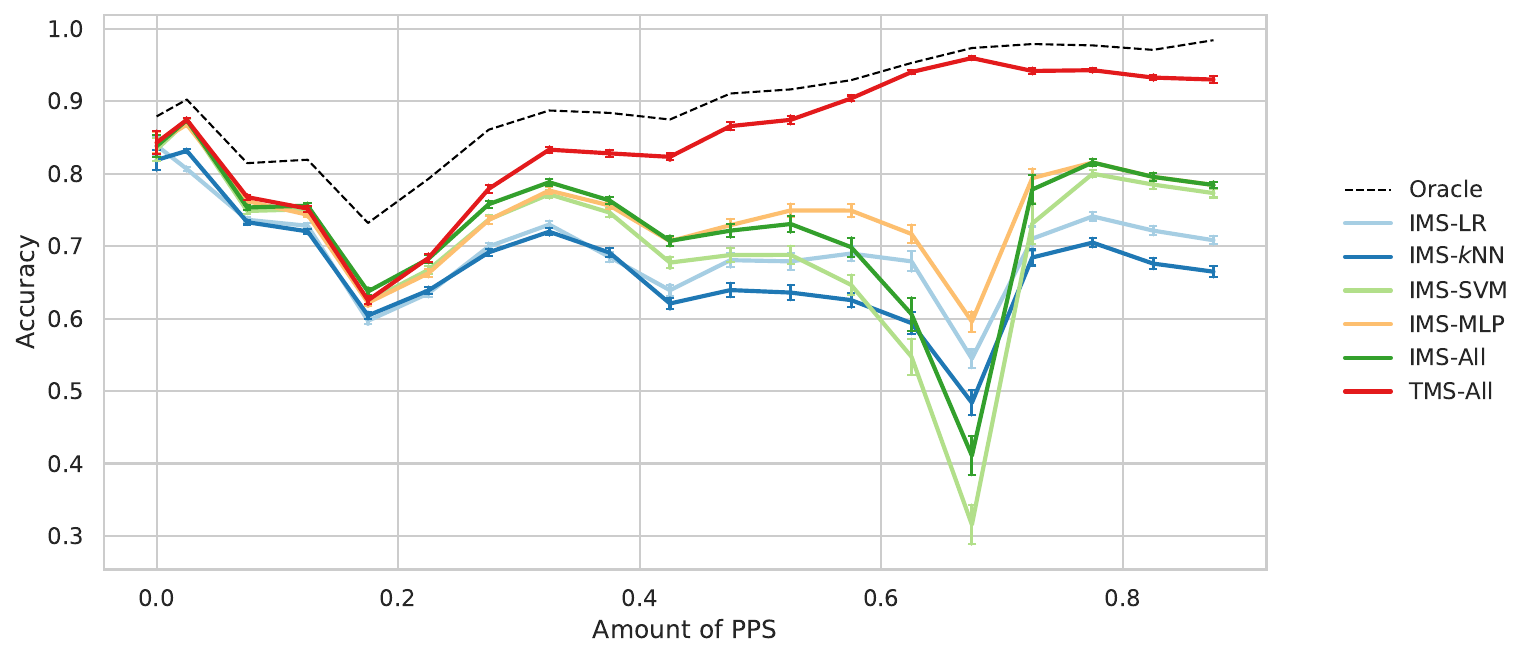}
 \caption{
 Accuracy of classifiers using different model selection strategies, as a function of the amount of PPS (measured in terms of the L1 distance between the training and test class prevalence distributions). Results are averaged across all test samples and datasets.
 %The amount of PPS is computed as the $L_1$ distance between the
% prevalence distribution of the training set $L$ and the prevalence
 %distribution of each test sample. 
 The black dashed line represents the selection oracle.
 %the average true accuracy of the \textit{true} best classifier in
 %each bin. 
 % \alexcomment{Ci aggiunerei la linea tratteggiata alla legend} 
 % \alexcomment{Aggiorna il discorso $\forall\rightarrow$All} 
 % \alexcomment{Magari farei il plot più lungo en meno alto, in modo da occupare tutto il textwidth e lasciare più spazio al testo} 
 % \alexcomment{ricorda pure di cambiare LEAP per TMS}
 % \alexcomment{Cambierei "True Accuracy" per "Accuracy" all'asse y}
 }
 \label{fig:shift}
\end{figure}

\noindent Table~\ref{tab:all_results}
% Tables~\ref{tab:results:bin} (for the binary case)
% and~\ref{tab:results:mul} (for the multiclass case)
reports the accuracy scores obtained by the classifiers resulting from
each of the model selection strategies considered; each accuracy value
is the average across the $\repetitions$ tests we have carried out for
that dataset.
% \lorenzocomment{Qui secondo me il discorso non è preciso: in realtà
% la media riportata è sull'accuracy dei 1000 classificatori scelti
% per i 1000 test samples (potenzialmente potrebbero essere tutti
% diversi). Da come diciamo mi sembra invece che prendiamo un
% classificatore per dataset e lo applichiamo ai 1000 test samples.}
% classifiers whose hyperparameters have been optimised either via the
% standard technique (IMS on the labelled data) or via TMS, averaged
% over the $\repetitions$ tests conducted for each dataset.
%
Overall, \tmsall\ tends to obtain the (per-dataset) best results,
% both in binary and in multiclass,
and obtains the greatest number of best results. In cases when
\tmsall\ does not obtain the best result, it still tends to obtain,
for most datasets, results that are not statistically significantly
different from the best-performing baseline.
% Even when it does not achieve the best result, \tmsall\ manages to
% still remain on a par with the best baselines in many cases.
%
Our experiments clearly show that adopting MS strategies tailored to
PPS yields a substantial performance advantage with respect to MS
techniques that assume IID data. Applying TMS appears also preferable
to training TSVMs; although TSVMs are tailored to transductive
contexts, in our experiments they underperform when facing scenarios
affected by PPS.

% While the performance of \tmsall\ appears to be consistent across
% most datasets, the na\"ive baselines show more variable
% results. Even \naive, which appears to be the best baseline overall,
% while showing good results, cannot match the overall performance of
% \tmsall; these results confirm what we have said in
% Section~\ref{sec:intro} about the fact that standard MS approaches
% do not take into account dataset shift.

% \fabseb{Argue that the plot of Figure~\ref{fig:shift} shows that,
% while our LEAP-ALL method and the standard IMS-ALL method are on a
% par for low levels of shift, LEAP-ALL drastically beats IMS-ALL for
% serious amounts of PPS (i.e., higher than 0.3).}

Figure~\ref{fig:shift} displays the (true) accuracy of the classifiers
as a function of the amount of PPS, measured as the L1 distance
between the vectors of class prevalence values of the training set and
test bag; the results are obtained as within-bin averages, where a bin
groups all the bags $U_i$ affected by a similar amount of PPS, across
all datasets.
A clear pattern emerging from the plot is that most methods perform
similarly for low levels of PPS, but when the intensity of the shift
increases there is a clear advantage in adopting TMS.

The same figure also displays (as a black dashed line) the performance
of an oracle, i.e., an idealized method that always picks the
best-performing classifier for each bag $U_i$.
The plot reveals that the difference between the oracle and
classifiers optimised via traditional MS strategies is relatively
small for low levels of shift (i.e., that IMS makes good choices in
near-IID scenarios). However, as the amount of shift increases,
traditional approaches suffer a substantial loss in performance; in
contrast, for these levels of shift the gap between the oracle and TMS
is fairly narrow, showing that in these contexts TMS performs very
well.

% It appears clear from the plot that, while our \tmsall\ method and
% the standard \imsall\ method are on a par for low levels of shift,
% \tmsall\ drastically beats \imsall\ for serious amounts of PPS
% (i.e., higher than 0.3). \lorenzocomment{ Also, notice the gap
% between each MS method and the perfect hyperparameter optimisation
% (represented by the black dashed line). For low levels of PPS this
% gap is relatively narrow. When the amount of PPS increases, the gap
% widens for all IMS methods, while remaining stable, or even getting
% narrower, for \tmsall. This indicates how increasing levels of PPS
% can impact on the performance of a traditional model selection
% approach, while confirming the robustness of TMS in this setting. }

% ------------------------------------------------------

\section{Conclusions}
\label{sec:conclusions}
\noindent We have discussed transductive model selection (TMS), a new
way of performing model selection (i.e., hyperparameter optimisation)
for classification applications in which the unlabelled data to be
classified are affected by dataset shift. Essentially, TMS replaces
traditional classifier accuracy
\textit{computation} %\alexcomment{Mi verrebbe da dire che questa computation, peró, altro non è che una stima del test (sbagliata under PPS, ma sempre una stima è)} \fabseb{Sì, ma gli iperparametri vengono ottimizzati calcolando l'accuratezza su dati labelled, il fatto importante è quello.}
on \textit{training} data with classifier accuracy \textit{estimation}
on the finite set of \textit{unlabelled} data that need to be
classified at a certain point in time; while the two would return the
same results in IID scenarios, they return different results in
situations characterised by dataset shift. We have presented TMS
experiments in a restricted setting, i.e., when the data are affected
by prior probability shift, an important type of dataset shift that
often affects anti-causal learning problems. Here, our experiments
have shown that TMS boosts classification accuracy, i.e., bring about
classifiers that outperform the classifiers whose hyperparameters have
been optimised, as usual, by cross-validation on the labelled data.

Note that TMS is not restricted to dealing with prior probability
shift, and can also deal with other types of shift too (e.g.,
covariate shift). For this, one only needs to use, in place of the
\oleap\ method used in this paper (which is tailored to prior
probability shift), a CAP method explicitly devised for the type of
shift that the data suffer from.

Aside from the application scenarios considered (i.e., those in which
the unlabelled data to be classified become available in successive
batches), TMS holds promise for all applications that have a strictly
transductive nature (i.e., in which \textit{all} the unlabelled data
to which the classifier needs to be applied are already known at
training time), such as technology-assisted review (TAR -- see
e.g.,~\cite{Gray:2024fv}) for supporting
e-discovery~\cite{Oard:2013fk}, online content
moderation~\cite{Yang:2021ks}, or the production of systematic
reviews~\cite{Ferdinands:2023hb}. Indeed, the next steps in our TMS
research will include its application to these domains.
% \fabseb{Le conclusioni mi sembrano a posto.}

% \alexcomment{Yet another related scenario regards
% \emph{technology-assisted review} (TAR) [citazione-chiedere
% Fabrizio], where the goal is to provide the most accurate label
% predictions for one specific test set, which is known and available
% at training time.}

% ------------------------------------------------------

%%
%% The acknowledgments section is defined using the "acknowledgments"
%% environment (and NOT an unnumbered section). This ensures the
%% proper identification of the section in the article metadata, and
%% the consistent spelling of the heading.
\section*{Acknowledgments}
 Lorenzo Volpi's work was supported by project ``Italian
 Strengthening of ESFRI RI RESILIENCE'' (ITSERR), funded by the
 European Union under the NextGenerationEU funding scheme (CUP
 B53C22001770006). Alejandro Moreo's and Fabrizio Sebastiani's work
 was partially supported by project ``Future Artificial Intelligence
 Research'' (FAIR), project ``Quantification under Dataset Shift''
 (QuaDaSh), and project ``Strengthening the Italian RI for Social
 Mining and Big Data Analytics'' (SoBigData.it), all funded by the
 European Union under the NextGenerationEU funding scheme (CUP
 B53D22000980006, CUP B53D23026250001, CUP B53C22001760006,
 respectively).

%% The declaration on generative AI comes in effect in Janary
%% 2025. See also https://ceur-ws.org/GenAI/Policy.html
%% \section*{Declaration on Generative AI}
%% {\em Either:}\newline The author(s) have not employed any
%% Generative AI tools. \newline
%% 
%% \noindent{\em Or (by using the activity taxonomy in ceur-ws.org/genai-tax.html):\newline}
%% During the preparation of this work, the author(s) used X-GPT-4 and
%% Gramby in order to: Grammar and spelling check. Further, the
%% author(s) used X-AI-IMG for figures 3 and 4 in order to: Generate
%% images. After using these tool(s)/service(s), the author(s)
%% reviewed and edited the content as needed and take(s) full
%% responsibility for the publication’s content.

%%
%% Define the bibliography file to be used
\bibliographystyle{apalike}
\bibliography{Fabrizio,Lorenzo}


% \appendix

\end{document}

%% file: tables/transd_vanilla_accuracy_table.tex
\begin{tabular}{|c|ccccc|ccccc|c|}
\cline{2-12}
\multicolumn{1}{c|}{} & \multicolumn{5}{c|}{$\varnothing$} & \multicolumn{5}{c|}{IMS} & \multicolumn{1}{c|}{TMS} \\\cline{2-12}
\multicolumn{1}{c|}{} & \nomslr & \nomsknn & \nomssvm & \nomsmlp & \nomstsvm & \imslr & \imsknn & \imssvm & \imsmlp & \imsall & \tmsall \\\hline
\textsf{poker-hand} & .121$^{\phantom{\dag}}\pm^{\phantom{\dag}}$.104\cellcolor{red!40} & .133$^{\phantom{\dag}}\pm^{\phantom{\dag}}$.080\cellcolor{red!19} & .139$^{\phantom{\dag}}\pm^{\phantom{\dag}}$.090\cellcolor{red!9} & \textbf{.169$^{\phantom{\dag}}\pm^{\phantom{\dag}}$.086}\cellcolor{green!40} & .137$^{\phantom{\dag}}\pm^{\phantom{\dag}}$.089\cellcolor{red!13} & .122$^{\phantom{\dag}}\pm^{\phantom{\dag}}$.109\cellcolor{red!38} & .133$^{\phantom{\dag}}\pm^{\phantom{\dag}}$.083\cellcolor{red!19} & .153$^{\phantom{\dag}}\pm^{\phantom{\dag}}$.082\cellcolor{green!13} & .150$^{\phantom{\dag}}\pm^{\phantom{\dag}}$.089\cellcolor{green!8} & .150$^{\phantom{\dag}}\pm^{\phantom{\dag}}$.089\cellcolor{green!8} & .134$^{\phantom{\dag}}\pm^{\phantom{\dag}}$.110\cellcolor{red!17} \\
\textsf{connect-4} & .515$^{\phantom{\dag}}\pm^{\phantom{\dag}}$.198\cellcolor{green!5} & .544$^{\phantom{\dag}}\pm^{\phantom{\dag}}$.146\cellcolor{green!12} & .539$^{\phantom{\dag}}\pm^{\phantom{\dag}}$.201\cellcolor{green!11} & .629$^{\phantom{\dag}}\pm^{\phantom{\dag}}$.136\cellcolor{green!33} & .334$^{\phantom{\dag}}\pm^{\phantom{\dag}}$.241\cellcolor{red!40} & .516$^{\phantom{\dag}}\pm^{\phantom{\dag}}$.198\cellcolor{green!5} & .512$^{\phantom{\dag}}\pm^{\phantom{\dag}}$.184\cellcolor{green!4} & .649$^{\phantom{\dag}}\pm^{\phantom{\dag}}$.114\cellcolor{green!38} & .632$^{\phantom{\dag}}\pm^{\phantom{\dag}}$.140\cellcolor{green!34} & .649$^{\phantom{\dag}}\pm^{\phantom{\dag}}$.114\cellcolor{green!38} & \textbf{.653$^{\phantom{\dag}}\pm^{\phantom{\dag}}$.142}\cellcolor{green!40} \\
\textsf{shuttle} & .716$^{\phantom{\dag}}\pm^{\phantom{\dag}}$.188\cellcolor{green!10} & .855$^{\phantom{\dag}}\pm^{\phantom{\dag}}$.114\cellcolor{green!32} & .790$^{\phantom{\dag}}\pm^{\phantom{\dag}}$.162\cellcolor{green!22} & .892$^{\phantom{\dag}}\pm^{\phantom{\dag}}$.087\cellcolor{green!38} & .409$^{\phantom{\dag}}\pm^{\phantom{\dag}}$.198\cellcolor{red!40} & .743$^{\phantom{\dag}}\pm^{\phantom{\dag}}$.190\cellcolor{green!14} & .866$^{\phantom{\dag}}\pm^{\phantom{\dag}}$.106\cellcolor{green!34} & \textbf{.900$^{\phantom{\dag}}\pm^{\phantom{\dag}}$.079}\cellcolor{green!40} & .879$^{\phantom{\dag}}\pm^{\phantom{\dag}}$.096\cellcolor{green!36} & .879$^{\phantom{\dag}}\pm^{\phantom{\dag}}$.096\cellcolor{green!36} & .880$^{\phantom{\dag}}\pm^{\phantom{\dag}}$.098\cellcolor{green!36} \\
\textsf{chess} & .340$^{\phantom{\dag}}\pm^{\phantom{\dag}}$.065\cellcolor{green!3} & .382$^{\phantom{\dag}}\pm^{\phantom{\dag}}$.061\cellcolor{green!10} & .409$^{\phantom{\dag}}\pm^{\phantom{\dag}}$.068\cellcolor{green!14} & .547$^{\phantom{\dag}}\pm^{\phantom{\dag}}$.068\cellcolor{green!37} & .074$^{\phantom{\dag}}\pm^{\phantom{\dag}}$.062\cellcolor{red!40} & .346$^{\phantom{\dag}}\pm^{\phantom{\dag}}$.066\cellcolor{green!4} & .449$^{\phantom{\dag}}\pm^{\phantom{\dag}}$.058\cellcolor{green!21} & \textbf{.564$^{\phantom{\dag}}\pm^{\phantom{\dag}}$.059}\cellcolor{green!40} & .547$^{\phantom{\dag}}\pm^{\phantom{\dag}}$.068\cellcolor{green!37} & \textbf{.564$^{\phantom{\dag}}\pm^{\phantom{\dag}}$.059}\cellcolor{green!40} & .561$^{\dag}\pm^{\phantom{\dag}}$.067\cellcolor{green!39} \\
\textsf{letter} & .763$^{\phantom{\dag}}\pm^{\phantom{\dag}}$.047\cellcolor{green!23} & .902$^{\phantom{\dag}}\pm^{\phantom{\dag}}$.031\cellcolor{green!35} & .911$^{\phantom{\dag}}\pm^{\phantom{\dag}}$.030\cellcolor{green!36} & .928$^{\phantom{\dag}}\pm^{\phantom{\dag}}$.027\cellcolor{green!38} & .055$^{\phantom{\dag}}\pm^{\phantom{\dag}}$.041\cellcolor{red!40} & .765$^{\phantom{\dag}}\pm^{\phantom{\dag}}$.047\cellcolor{green!23} & .915$^{\phantom{\dag}}\pm^{\phantom{\dag}}$.029\cellcolor{green!37} & \textbf{.948$^{\phantom{\dag}}\pm^{\phantom{\dag}}$.023}\cellcolor{green!40} & .928$^{\phantom{\dag}}\pm^{\phantom{\dag}}$.027\cellcolor{green!38} & \textbf{.948$^{\phantom{\dag}}\pm^{\phantom{\dag}}$.023}\cellcolor{green!40} & .947$^{\dag}\pm^{\phantom{\dag}}$.024\cellcolor{green!39} \\
\textsf{dry-bean} & .936$^{\phantom{\dag}}\pm^{\phantom{\dag}}$.027\cellcolor{green!39} & .934$^{\phantom{\dag}}\pm^{\phantom{\dag}}$.028\cellcolor{green!39} & .938$^{\phantom{\dag}}\pm^{\phantom{\dag}}$.027\cellcolor{green!39} & \textbf{.940$^{\phantom{\dag}}\pm^{\phantom{\dag}}$.027}\cellcolor{green!40} & .250$^{\phantom{\dag}}\pm^{\phantom{\dag}}$.141\cellcolor{red!40} & .936$^{\phantom{\dag}}\pm^{\phantom{\dag}}$.027\cellcolor{green!39} & .932$^{\phantom{\dag}}\pm^{\phantom{\dag}}$.028\cellcolor{green!39} & .940$^{\dag}\pm^{\phantom{\dag}}$.027\cellcolor{green!39} & .938$^{\phantom{\dag}}\pm^{\phantom{\dag}}$.027\cellcolor{green!39} & .938$^{\phantom{\dag}}\pm^{\phantom{\dag}}$.027\cellcolor{green!39} & .938$^{\phantom{\dag}}\pm^{\phantom{\dag}}$.073\cellcolor{green!39} \\
\textsf{nursery} & .850$^{\phantom{\dag}}\pm^{\phantom{\dag}}$.079\cellcolor{green!24} & .715$^{\phantom{\dag}}\pm^{\phantom{\dag}}$.139\cellcolor{green!9} & .868$^{\phantom{\dag}}\pm^{\phantom{\dag}}$.105\cellcolor{green!26} & .997$^{\phantom{\dag}}\pm^{\phantom{\dag}}$.006\cellcolor{green!39} & .249$^{\phantom{\dag}}\pm^{\phantom{\dag}}$.194\cellcolor{red!40} & .876$^{\phantom{\dag}}\pm^{\phantom{\dag}}$.065\cellcolor{green!27} & .743$^{\phantom{\dag}}\pm^{\phantom{\dag}}$.156\cellcolor{green!12} & \textbf{.997$^{\phantom{\dag}}\pm^{\phantom{\dag}}$.006}\cellcolor{green!40} & .993$^{\phantom{\dag}}\pm^{\phantom{\dag}}$.010\cellcolor{green!39} & \textbf{.997$^{\phantom{\dag}}\pm^{\phantom{\dag}}$.006}\cellcolor{green!40} & .996$^{\phantom{\dag}}\pm^{\phantom{\dag}}$.007\cellcolor{green!39} \\
\textsf{hand-digits} & .940$^{\phantom{\dag}}\pm^{\phantom{\dag}}$.026\cellcolor{green!34} & .988$^{\phantom{\dag}}\pm^{\phantom{\dag}}$.011\cellcolor{green!39} & .992$^{\phantom{\dag}}\pm^{\phantom{\dag}}$.009\cellcolor{green!39} & .992$^{\phantom{\dag}}\pm^{\phantom{\dag}}$.009\cellcolor{green!39} & .182$^{\phantom{\dag}}\pm^{\phantom{\dag}}$.112\cellcolor{red!40} & .944$^{\phantom{\dag}}\pm^{\phantom{\dag}}$.025\cellcolor{green!35} & .989$^{\phantom{\dag}}\pm^{\phantom{\dag}}$.011\cellcolor{green!39} & \textbf{.995$^{\phantom{\dag}}\pm^{\phantom{\dag}}$.007}\cellcolor{green!40} & .992$^{\phantom{\dag}}\pm^{\phantom{\dag}}$.009\cellcolor{green!39} & \textbf{.995$^{\phantom{\dag}}\pm^{\phantom{\dag}}$.007}\cellcolor{green!40} & .994$^{\phantom{\dag}}\pm^{\phantom{\dag}}$.008\cellcolor{green!39} \\
\textsf{isolet} & .952$^{\phantom{\dag}}\pm^{\phantom{\dag}}$.022\cellcolor{green!39} & .856$^{\phantom{\dag}}\pm^{\phantom{\dag}}$.042\cellcolor{green!30} & .954$^{\phantom{\dag}}\pm^{\phantom{\dag}}$.022\cellcolor{green!39} & .949$^{\phantom{\dag}}\pm^{\phantom{\dag}}$.024\cellcolor{green!38} & .036$^{\phantom{\dag}}\pm^{\phantom{\dag}}$.034\cellcolor{red!40} & .951$^{\phantom{\dag}}\pm^{\phantom{\dag}}$.022\cellcolor{green!39} & .882$^{\phantom{\dag}}\pm^{\phantom{\dag}}$.038\cellcolor{green!33} & \textbf{.960$^{\phantom{\dag}}\pm^{\phantom{\dag}}$.020}\cellcolor{green!40} & .951$^{\phantom{\dag}}\pm^{\phantom{\dag}}$.023\cellcolor{green!39} & \textbf{.960$^{\phantom{\dag}}\pm^{\phantom{\dag}}$.020}\cellcolor{green!40} & .960$^{\dag}\pm^{\phantom{\dag}}$.021\cellcolor{green!39} \\
\textsf{wine-quality} & .306$^{\phantom{\dag}}\pm^{\phantom{\dag}}$.124\cellcolor{red!2} & .341$^{\phantom{\dag}}\pm^{\phantom{\dag}}$.102\cellcolor{green!10} & .306$^{\phantom{\dag}}\pm^{\phantom{\dag}}$.130\cellcolor{red!2} & .329$^{\phantom{\dag}}\pm^{\phantom{\dag}}$.121\cellcolor{green!6} & .200$^{\phantom{\dag}}\pm^{\phantom{\dag}}$.166\cellcolor{red!40} & .313$^{\phantom{\dag}}\pm^{\phantom{\dag}}$.121\cellcolor{green!0} & \textbf{.423$^{\phantom{\dag}}\pm^{\phantom{\dag}}$.102}\cellcolor{green!40} & .354$^{\phantom{\dag}}\pm^{\phantom{\dag}}$.127\cellcolor{green!15} & .319$^{\phantom{\dag}}\pm^{\phantom{\dag}}$.122\cellcolor{green!2} & \textbf{.423$^{\phantom{\dag}}\pm^{\phantom{\dag}}$.102}\cellcolor{green!40} & .349$^{\phantom{\dag}}\pm^{\phantom{\dag}}$.154\cellcolor{green!13} \\
\textsf{satellite} & .807$^{\phantom{\dag}}\pm^{\phantom{\dag}}$.079\cellcolor{green!29} & .870$^{\phantom{\dag}}\pm^{\phantom{\dag}}$.048\cellcolor{green!37} & .863$^{\phantom{\dag}}\pm^{\phantom{\dag}}$.057\cellcolor{green!36} & .882$^{\phantom{\dag}}\pm^{\phantom{\dag}}$.050\cellcolor{green!39} & .257$^{\phantom{\dag}}\pm^{\phantom{\dag}}$.143\cellcolor{red!40} & .806$^{\phantom{\dag}}\pm^{\phantom{\dag}}$.077\cellcolor{green!29} & .874$^{\phantom{\dag}}\pm^{\phantom{\dag}}$.047\cellcolor{green!38} & .882$^{\phantom{\dag}}\pm^{\phantom{\dag}}$.052\cellcolor{green!39} & .877$^{\phantom{\dag}}\pm^{\phantom{\dag}}$.053\cellcolor{green!38} & .882$^{\phantom{\dag}}\pm^{\phantom{\dag}}$.052\cellcolor{green!39} & \textbf{.888$^{\phantom{\dag}}\pm^{\phantom{\dag}}$.048}\cellcolor{green!40} \\
\textsf{digits} & .963$^{\phantom{\dag}}\pm^{\phantom{\dag}}$.020\cellcolor{green!38} & .961$^{\phantom{\dag}}\pm^{\phantom{\dag}}$.021\cellcolor{green!38} & .973$^{\phantom{\dag}}\pm^{\phantom{\dag}}$.017\cellcolor{green!39} & .972$^{\phantom{\dag}}\pm^{\phantom{\dag}}$.018\cellcolor{green!39} & .100$^{\phantom{\dag}}\pm^{\phantom{\dag}}$.089\cellcolor{red!40} & .962$^{\phantom{\dag}}\pm^{\phantom{\dag}}$.020\cellcolor{green!38} & .962$^{\phantom{\dag}}\pm^{\phantom{\dag}}$.021\cellcolor{green!38} & \textbf{.977$^{\phantom{\dag}}\pm^{\phantom{\dag}}$.016}\cellcolor{green!40} & .974$^{\phantom{\dag}}\pm^{\phantom{\dag}}$.017\cellcolor{green!39} & \textbf{.977$^{\phantom{\dag}}\pm^{\phantom{\dag}}$.016}\cellcolor{green!40} & .976$^{\dag}\pm^{\phantom{\dag}}$.016\cellcolor{green!39} \\
\textsf{page-block} & .713$^{\phantom{\dag}}\pm^{\phantom{\dag}}$.135\cellcolor{red!1} & .703$^{\phantom{\dag}}\pm^{\phantom{\dag}}$.158\cellcolor{red!3} & .666$^{\phantom{\dag}}\pm^{\phantom{\dag}}$.168\cellcolor{red!11} & .808$^{\phantom{\dag}}\pm^{\phantom{\dag}}$.100\cellcolor{green!19} & .537$^{\phantom{\dag}}\pm^{\phantom{\dag}}$.214\cellcolor{red!40} & .758$^{\phantom{\dag}}\pm^{\phantom{\dag}}$.126\cellcolor{green!8} & .716$^{\phantom{\dag}}\pm^{\phantom{\dag}}$.152\cellcolor{red!0} & .799$^{\phantom{\dag}}\pm^{\phantom{\dag}}$.100\cellcolor{green!17} & .850$^{\phantom{\dag}}\pm^{\phantom{\dag}}$.075\cellcolor{green!28} & .850$^{\phantom{\dag}}\pm^{\phantom{\dag}}$.075\cellcolor{green!28} & \textbf{.903$^{\phantom{\dag}}\pm^{\phantom{\dag}}$.038}\cellcolor{green!40} \\
\textsf{waveform-v1} & .867$^{\phantom{\dag}}\pm^{\phantom{\dag}}$.038\cellcolor{green!39} & .813$^{\phantom{\dag}}\pm^{\phantom{\dag}}$.048\cellcolor{green!31} & .858$^{\phantom{\dag}}\pm^{\phantom{\dag}}$.039\cellcolor{green!38} & .844$^{\phantom{\dag}}\pm^{\phantom{\dag}}$.037\cellcolor{green!35} & .334$^{\phantom{\dag}}\pm^{\phantom{\dag}}$.241\cellcolor{red!40} & .866$^{\phantom{\dag}}\pm^{\phantom{\dag}}$.040\cellcolor{green!39} & .843$^{\phantom{\dag}}\pm^{\phantom{\dag}}$.049\cellcolor{green!35} & .861$^{\phantom{\dag}}\pm^{\phantom{\dag}}$.056\cellcolor{green!38} & .844$^{\phantom{\dag}}\pm^{\phantom{\dag}}$.038\cellcolor{green!35} & .866$^{\phantom{\dag}}\pm^{\phantom{\dag}}$.040\cellcolor{green!39} & \textbf{.871$^{\phantom{\dag}}\pm^{\phantom{\dag}}$.101}\cellcolor{green!40} \\
\textsf{spambase} & .922$^{\phantom{\dag}}\pm^{\phantom{\dag}}$.030\cellcolor{green!35} & .877$^{\phantom{\dag}}\pm^{\phantom{\dag}}$.042\cellcolor{green!25} & .924$^{\phantom{\dag}}\pm^{\phantom{\dag}}$.034\cellcolor{green!35} & .937$^{\phantom{\dag}}\pm^{\phantom{\dag}}$.025\cellcolor{green!38} & .580$^{\phantom{\dag}}\pm^{\phantom{\dag}}$.247\cellcolor{red!40} & .925$^{\phantom{\dag}}\pm^{\phantom{\dag}}$.028\cellcolor{green!35} & .893$^{\phantom{\dag}}\pm^{\phantom{\dag}}$.038\cellcolor{green!28} & .915$^{\phantom{\dag}}\pm^{\phantom{\dag}}$.039\cellcolor{green!33} & .935$^{\phantom{\dag}}\pm^{\phantom{\dag}}$.028\cellcolor{green!37} & .935$^{\phantom{\dag}}\pm^{\phantom{\dag}}$.028\cellcolor{green!37} & \textbf{.944$^{\phantom{\dag}}\pm^{\phantom{\dag}}$.027}\cellcolor{green!40} \\
\textsf{academic-success} & .652$^{\phantom{\dag}}\pm^{\phantom{\dag}}$.150\cellcolor{green!23} & .555$^{\phantom{\dag}}\pm^{\phantom{\dag}}$.147\cellcolor{green!4} & .631$^{\phantom{\dag}}\pm^{\phantom{\dag}}$.151\cellcolor{green!19} & .668$^{\phantom{\dag}}\pm^{\phantom{\dag}}$.116\cellcolor{green!26} & .334$^{\phantom{\dag}}\pm^{\phantom{\dag}}$.241\cellcolor{red!40} & .652$^{\phantom{\dag}}\pm^{\phantom{\dag}}$.150\cellcolor{green!23} & .561$^{\phantom{\dag}}\pm^{\phantom{\dag}}$.180\cellcolor{green!5} & .669$^{\phantom{\dag}}\pm^{\phantom{\dag}}$.055\cellcolor{green!26} & .656$^{\phantom{\dag}}\pm^{\phantom{\dag}}$.118\cellcolor{green!24} & .669$^{\phantom{\dag}}\pm^{\phantom{\dag}}$.055\cellcolor{green!26} & \textbf{.736$^{\phantom{\dag}}\pm^{\phantom{\dag}}$.096}\cellcolor{green!40} \\
\textsf{abalone} & .255$^{\phantom{\dag}}\pm^{\phantom{\dag}}$.069\cellcolor{green!37} & .231$^{\phantom{\dag}}\pm^{\phantom{\dag}}$.059\cellcolor{green!26} & .249$^{\phantom{\dag}}\pm^{\phantom{\dag}}$.071\cellcolor{green!34} & .255$^{\phantom{\dag}}\pm^{\phantom{\dag}}$.066\cellcolor{green!37} & .108$^{\phantom{\dag}}\pm^{\phantom{\dag}}$.079\cellcolor{red!31} & .255$^{\phantom{\dag}}\pm^{\phantom{\dag}}$.069\cellcolor{green!37} & .248$^{\phantom{\dag}}\pm^{\phantom{\dag}}$.067\cellcolor{green!34} & .090$^{\phantom{\dag}}\pm^{\phantom{\dag}}$.083\cellcolor{red!40} & \textbf{.260$^{\phantom{\dag}}\pm^{\phantom{\dag}}$.066}\cellcolor{green!40} & .090$^{\phantom{\dag}}\pm^{\phantom{\dag}}$.083\cellcolor{red!40} & .233$^{\phantom{\dag}}\pm^{\phantom{\dag}}$.104\cellcolor{green!26} \\
\textsf{molecular} & .903$^{\phantom{\dag}}\pm^{\phantom{\dag}}$.030\cellcolor{green!34} & .817$^{\phantom{\dag}}\pm^{\phantom{\dag}}$.060\cellcolor{green!22} & .934$^{\phantom{\dag}}\pm^{\phantom{\dag}}$.027\cellcolor{green!38} & .921$^{\phantom{\dag}}\pm^{\phantom{\dag}}$.028\cellcolor{green!36} & .340$^{\phantom{\dag}}\pm^{\phantom{\dag}}$.239\cellcolor{red!40} & .934$^{\phantom{\dag}}\pm^{\phantom{\dag}}$.025\cellcolor{green!38} & .872$^{\phantom{\dag}}\pm^{\phantom{\dag}}$.042\cellcolor{green!30} & .938$^{\phantom{\dag}}\pm^{\phantom{\dag}}$.029\cellcolor{green!38} & .933$^{\phantom{\dag}}\pm^{\phantom{\dag}}$.027\cellcolor{green!38} & .938$^{\phantom{\dag}}\pm^{\phantom{\dag}}$.029\cellcolor{green!38} & \textbf{.946$^{\phantom{\dag}}\pm^{\phantom{\dag}}$.039}\cellcolor{green!40} \\
\textsf{image-seg} & .931$^{\phantom{\dag}}\pm^{\phantom{\dag}}$.038\cellcolor{green!37} & .905$^{\phantom{\dag}}\pm^{\phantom{\dag}}$.050\cellcolor{green!34} & .918$^{\phantom{\dag}}\pm^{\phantom{\dag}}$.048\cellcolor{green!36} & .928$^{\phantom{\dag}}\pm^{\phantom{\dag}}$.044\cellcolor{green!37} & .277$^{\phantom{\dag}}\pm^{\phantom{\dag}}$.154\cellcolor{red!40} & .947$^{\phantom{\dag}}\pm^{\phantom{\dag}}$.032\cellcolor{green!39} & .914$^{\phantom{\dag}}\pm^{\phantom{\dag}}$.045\cellcolor{green!35} & .945$^{\phantom{\dag}}\pm^{\phantom{\dag}}$.030\cellcolor{green!39} & .934$^{\phantom{\dag}}\pm^{\phantom{\dag}}$.041\cellcolor{green!38} & .945$^{\phantom{\dag}}\pm^{\phantom{\dag}}$.030\cellcolor{green!39} & \textbf{.951$^{\phantom{\dag}}\pm^{\phantom{\dag}}$.029}\cellcolor{green!40} \\
\textsf{obesity} & .821$^{\phantom{\dag}}\pm^{\phantom{\dag}}$.064\cellcolor{green!28} & .718$^{\phantom{\dag}}\pm^{\phantom{\dag}}$.080\cellcolor{green!17} & .811$^{\phantom{\dag}}\pm^{\phantom{\dag}}$.057\cellcolor{green!27} & .846$^{\phantom{\dag}}\pm^{\phantom{\dag}}$.052\cellcolor{green!31} & .163$^{\phantom{\dag}}\pm^{\phantom{\dag}}$.119\cellcolor{red!40} & \textbf{.931$^{\phantom{\dag}}\pm^{\phantom{\dag}}$.032}\cellcolor{green!40} & .753$^{\phantom{\dag}}\pm^{\phantom{\dag}}$.079\cellcolor{green!21} & .894$^{\phantom{\dag}}\pm^{\phantom{\dag}}$.041\cellcolor{green!36} & .846$^{\phantom{\dag}}\pm^{\phantom{\dag}}$.052\cellcolor{green!31} & \textbf{.931$^{\phantom{\dag}}\pm^{\phantom{\dag}}$.032}\cellcolor{green!40} & .924$^{\phantom{\dag}}\pm^{\phantom{\dag}}$.034\cellcolor{green!39} \\
\textsf{semeion} & .848$^{\phantom{\dag}}\pm^{\phantom{\dag}}$.090\cellcolor{green!25} & .872$^{\phantom{\dag}}\pm^{\phantom{\dag}}$.080\cellcolor{green!30} & .830$^{\phantom{\dag}}\pm^{\phantom{\dag}}$.106\cellcolor{green!22} & .914$^{\phantom{\dag}}\pm^{\phantom{\dag}}$.056\cellcolor{green!38} & .504$^{\phantom{\dag}}\pm^{\phantom{\dag}}$.291\cellcolor{red!40} & .848$^{\phantom{\dag}}\pm^{\phantom{\dag}}$.090\cellcolor{green!25} & .829$^{\phantom{\dag}}\pm^{\phantom{\dag}}$.105\cellcolor{green!22} & .884$^{\phantom{\dag}}\pm^{\phantom{\dag}}$.074\cellcolor{green!32} & .891$^{\phantom{\dag}}\pm^{\phantom{\dag}}$.068\cellcolor{green!34} & .891$^{\phantom{\dag}}\pm^{\phantom{\dag}}$.068\cellcolor{green!34} & \textbf{.922$^{\phantom{\dag}}\pm^{\phantom{\dag}}$.106}\cellcolor{green!40} \\
\textsf{yeast} & .613$^{\phantom{\dag}}\pm^{\phantom{\dag}}$.090\cellcolor{green!13} & .635$^{\phantom{\dag}}\pm^{\phantom{\dag}}$.083\cellcolor{green!17} & .648$^{\phantom{\dag}}\pm^{\phantom{\dag}}$.083\cellcolor{green!19} & .646$^{\phantom{\dag}}\pm^{\phantom{\dag}}$.077\cellcolor{green!19} & .324$^{\phantom{\dag}}\pm^{\phantom{\dag}}$.185\cellcolor{red!40} & .641$^{\phantom{\dag}}\pm^{\phantom{\dag}}$.129\cellcolor{green!18} & .654$^{\phantom{\dag}}\pm^{\phantom{\dag}}$.097\cellcolor{green!21} & .581$^{\phantom{\dag}}\pm^{\phantom{\dag}}$.227\cellcolor{green!7} & .679$^{\phantom{\dag}}\pm^{\phantom{\dag}}$.095\cellcolor{green!25} & .641$^{\phantom{\dag}}\pm^{\phantom{\dag}}$.129\cellcolor{green!18} & \textbf{.756$^{\phantom{\dag}}\pm^{\phantom{\dag}}$.178}\cellcolor{green!40} \\
\textsf{cmc} & .483$^{\phantom{\dag}}\pm^{\phantom{\dag}}$.076\cellcolor{green!14} & .441$^{\phantom{\dag}}\pm^{\phantom{\dag}}$.068\cellcolor{red!1} & .494$^{\phantom{\dag}}\pm^{\phantom{\dag}}$.069\cellcolor{green!19} & .469$^{\phantom{\dag}}\pm^{\phantom{\dag}}$.076\cellcolor{green!9} & .344$^{\phantom{\dag}}\pm^{\phantom{\dag}}$.222\cellcolor{red!40} & .526$^{\phantom{\dag}}\pm^{\phantom{\dag}}$.077\cellcolor{green!31} & .441$^{\phantom{\dag}}\pm^{\phantom{\dag}}$.068\cellcolor{red!1} & .354$^{\phantom{\dag}}\pm^{\phantom{\dag}}$.229\cellcolor{red!36} & .475$^{\phantom{\dag}}\pm^{\phantom{\dag}}$.074\cellcolor{green!11} & .475$^{\phantom{\dag}}\pm^{\phantom{\dag}}$.074\cellcolor{green!11} & \textbf{.547$^{\phantom{\dag}}\pm^{\phantom{\dag}}$.198}\cellcolor{green!40} \\
\textsf{hcv} & .251$^{\phantom{\dag}}\pm^{\phantom{\dag}}$.049\cellcolor{green!13} & .251$^{\phantom{\dag}}\pm^{\phantom{\dag}}$.048\cellcolor{green!15} & .239$^{\phantom{\dag}}\pm^{\phantom{\dag}}$.055\cellcolor{red!15} & .246$^{\phantom{\dag}}\pm^{\phantom{\dag}}$.051\cellcolor{green!3} & .249$^{\phantom{\dag}}\pm^{\phantom{\dag}}$.194\cellcolor{green!8} & .229$^{\phantom{\dag}}\pm^{\phantom{\dag}}$.045\cellcolor{red!40} & \textbf{.261$^{\phantom{\dag}}\pm^{\phantom{\dag}}$.052}\cellcolor{green!40} & .248$^{\phantom{\dag}}\pm^{\phantom{\dag}}$.189\cellcolor{green!7} & .236$^{\phantom{\dag}}\pm^{\phantom{\dag}}$.053\cellcolor{red!21} & .248$^{\phantom{\dag}}\pm^{\phantom{\dag}}$.189\cellcolor{green!7} & .238$^{\phantom{\dag}}\pm^{\phantom{\dag}}$.188\cellcolor{red!16} \\
\textsf{phishing} & .635$^{\phantom{\dag}}\pm^{\phantom{\dag}}$.196\cellcolor{red!8} & .682$^{\phantom{\dag}}\pm^{\phantom{\dag}}$.146\cellcolor{green!4} & .623$^{\phantom{\dag}}\pm^{\phantom{\dag}}$.215\cellcolor{red!12} & .704$^{\phantom{\dag}}\pm^{\phantom{\dag}}$.153\cellcolor{green!10} & .525$^{\phantom{\dag}}\pm^{\phantom{\dag}}$.199\cellcolor{red!40} & .635$^{\phantom{\dag}}\pm^{\phantom{\dag}}$.196\cellcolor{red!8} & .725$^{\phantom{\dag}}\pm^{\phantom{\dag}}$.109\cellcolor{green!16} & \textbf{.806$^{\phantom{\dag}}\pm^{\phantom{\dag}}$.072}\cellcolor{green!40} & .715$^{\phantom{\dag}}\pm^{\phantom{\dag}}$.146\cellcolor{green!14} & \textbf{.806$^{\phantom{\dag}}\pm^{\phantom{\dag}}$.072}\cellcolor{green!40} & .743$^{\phantom{\dag}}\pm^{\phantom{\dag}}$.173\cellcolor{green!21} \\
\textsf{mhr} & .629$^{\phantom{\dag}}\pm^{\phantom{\dag}}$.113\cellcolor{red!21} & .714$^{\phantom{\dag}}\pm^{\phantom{\dag}}$.087\cellcolor{green!7} & .721$^{\phantom{\dag}}\pm^{\phantom{\dag}}$.117\cellcolor{green!9} & .701$^{\phantom{\dag}}\pm^{\phantom{\dag}}$.099\cellcolor{green!3} & .574$^{\phantom{\dag}}\pm^{\phantom{\dag}}$.217\cellcolor{red!40} & .651$^{\phantom{\dag}}\pm^{\phantom{\dag}}$.153\cellcolor{red!13} & \textbf{.810$^{\phantom{\dag}}\pm^{\phantom{\dag}}$.044}\cellcolor{green!40} & .729$^{\phantom{\dag}}\pm^{\phantom{\dag}}$.070\cellcolor{green!12} & .731$^{\phantom{\dag}}\pm^{\phantom{\dag}}$.095\cellcolor{green!13} & \textbf{.810$^{\phantom{\dag}}\pm^{\phantom{\dag}}$.044}\cellcolor{green!40} & .795$^{\dag}\pm^{\phantom{\dag}}$.110\cellcolor{green!34} \\
\textsf{german} & .657$^{\phantom{\dag}}\pm^{\phantom{\dag}}$.128\cellcolor{green!7} & .604$^{\phantom{\dag}}\pm^{\phantom{\dag}}$.150\cellcolor{red!8} & .610$^{\phantom{\dag}}\pm^{\phantom{\dag}}$.196\cellcolor{red!6} & .668$^{\phantom{\dag}}\pm^{\phantom{\dag}}$.106\cellcolor{green!10} & .496$^{\phantom{\dag}}\pm^{\phantom{\dag}}$.291\cellcolor{red!40} & .669$^{\phantom{\dag}}\pm^{\phantom{\dag}}$.100\cellcolor{green!10} & .574$^{\phantom{\dag}}\pm^{\phantom{\dag}}$.216\cellcolor{red!17} & .496$^{\phantom{\dag}}\pm^{\phantom{\dag}}$.291\cellcolor{red!40} & .642$^{\phantom{\dag}}\pm^{\phantom{\dag}}$.128\cellcolor{green!2} & .496$^{\phantom{\dag}}\pm^{\phantom{\dag}}$.291\cellcolor{red!40} & \textbf{.769$^{\phantom{\dag}}\pm^{\phantom{\dag}}$.137}\cellcolor{green!40} \\
\textsf{tictactoe} & .975$^{\phantom{\dag}}\pm^{\phantom{\dag}}$.021\cellcolor{green!38} & .900$^{\phantom{\dag}}\pm^{\phantom{\dag}}$.054\cellcolor{green!25} & .960$^{\phantom{\dag}}\pm^{\phantom{\dag}}$.030\cellcolor{green!35} & .970$^{\phantom{\dag}}\pm^{\phantom{\dag}}$.024\cellcolor{green!37} & .504$^{\phantom{\dag}}\pm^{\phantom{\dag}}$.291\cellcolor{red!40} & .980$^{\phantom{\dag}}\pm^{\phantom{\dag}}$.018\cellcolor{green!39} & .900$^{\phantom{\dag}}\pm^{\phantom{\dag}}$.054\cellcolor{green!25} & \textbf{.985$^{\phantom{\dag}}\pm^{\phantom{\dag}}$.015}\cellcolor{green!40} & .970$^{\phantom{\dag}}\pm^{\phantom{\dag}}$.024\cellcolor{green!37} & \textbf{.985$^{\phantom{\dag}}\pm^{\phantom{\dag}}$.015}\cellcolor{green!40} & .981$^{\phantom{\dag}}\pm^{\phantom{\dag}}$.044\cellcolor{green!39} \\
\textsf{mammographic} & .788$^{\phantom{\dag}}\pm^{\phantom{\dag}}$.044\cellcolor{green!0} & .760$^{\phantom{\dag}}\pm^{\phantom{\dag}}$.043\cellcolor{red!22} & .760$^{\phantom{\dag}}\pm^{\phantom{\dag}}$.051\cellcolor{red!23} & .792$^{\phantom{\dag}}\pm^{\phantom{\dag}}$.049\cellcolor{green!2} & .739$^{\phantom{\dag}}\pm^{\phantom{\dag}}$.046\cellcolor{red!40} & .800$^{\phantom{\dag}}\pm^{\phantom{\dag}}$.044\cellcolor{green!9} & .769$^{\phantom{\dag}}\pm^{\phantom{\dag}}$.049\cellcolor{red!15} & .787$^{\phantom{\dag}}\pm^{\phantom{\dag}}$.046\cellcolor{red!0} & .784$^{\phantom{\dag}}\pm^{\phantom{\dag}}$.045\cellcolor{red!3} & .784$^{\phantom{\dag}}\pm^{\phantom{\dag}}$.045\cellcolor{red!3} & \textbf{.837$^{\phantom{\dag}}\pm^{\phantom{\dag}}$.059}\cellcolor{green!40} \\\hline
\textit{Average} & .695$^{\phantom{\dag}}\pm^{\phantom{\dag}}$.261\cellcolor{green!26} & .688$^{\phantom{\dag}}\pm^{\phantom{\dag}}$.250\cellcolor{green!25} & .709$^{\phantom{\dag}}\pm^{\phantom{\dag}}$.265\cellcolor{green!28} & .743$^{\phantom{\dag}}\pm^{\phantom{\dag}}$.252\cellcolor{green!34} & .318$^{\phantom{\dag}}\pm^{\phantom{\dag}}$.259\cellcolor{red!40} & .706$^{\phantom{\dag}}\pm^{\phantom{\dag}}$.262\cellcolor{green!28} & .703$^{\phantom{\dag}}\pm^{\phantom{\dag}}$.247\cellcolor{green!27} & .734$^{\phantom{\dag}}\pm^{\phantom{\dag}}$.285\cellcolor{green!33} & .741$^{\phantom{\dag}}\pm^{\phantom{\dag}}$.251\cellcolor{green!34} & .745$^{\phantom{\dag}}\pm^{\phantom{\dag}}$.270\cellcolor{green!35} & \textbf{.771$^{\phantom{\dag}}\pm^{\phantom{\dag}}$.265}\cellcolor{green!40} \\\hline
\end{tabular}